\documentclass[runningheads]{llncs}
\usepackage[T1]{fontenc}
\usepackage{graphicx}

\usepackage{epstopdf}
\usepackage{comment}
\usepackage{lineno,hyperref}
\usepackage{caption}
\usepackage{subfigure}
\usepackage{stfloats}
\usepackage{booktabs}
\usepackage{threeparttable}
\usepackage{multicol}
\usepackage{multirow}
\usepackage{enumerate}
\usepackage{fancyhdr}
\usepackage{amsmath}
\usepackage{cases}
\usepackage{lastpage}
\usepackage{algorithm}
\usepackage{algpseudocode}
\usepackage{hyperref}
\usepackage{textcomp}
\usepackage{amssymb}
\usepackage{color}
\usepackage{marvosym}
\usepackage{gensymb}
\usepackage[marginal]{footmisc}

\begin{document}

\title{Landslide Surface Displacement Prediction Based on VSXC-LSTM Algorithm}

\author{
Menglin Kong\inst{1}
\and Ruichen Li\inst{1}
\and Fan Liu\inst{1}
\and Xingquan Li\inst{2} \inst{3}
\and Juan Cheng\inst{1}
\and Muzhou Hou\inst{1}
\and Cong Cao\inst{1}{\textsuperscript{\Letter}}
\thanks{Menglin Kong and Ruichen Li contributed equally to this work. Cong Cao is the corresponding author(congcao@csu.edu.cn).}
}
\authorrunning{Kong et al.}
\institute{School of Mathematics and Statistics, Central South University, Changsha, China\\
\and Peng Cheng Laboratory, Shenzhen, China\\
\and School of Mathematics and Statistics, Minnan Normal University, Zhangzhou, China\\
\email{\{212112025,212111114,8201200822,hmzw,congcao\}@csu.edu.cn,chengjuan0306@163.com,fzulxq@gmail.com}}


\maketitle

\begin{abstract}
Landslide is a natural disaster that can easily threaten local ecology, people's lives and property. In this paper, we conduct modelling research on real unidirectional surface displacement data of recent landslides in the research area and propose a time series prediction framework named VMD-SegSigmoid-XGBoost-ClusterLSTM (VSXC-LSTM) based on variational mode decomposition, which can predict the landslide surface displacement more accurately. The model performs well on the test set. Except for the random item subsequence that is hard to fit, the root mean square error (RMSE) and the mean absolute percentage error (MAPE) of the trend item subsequence and the periodic item subsequence are both less than 0.1, and the RMSE is as low as 0.006 for the periodic item prediction module based on XGBoost\footnote{Accepted in ICANN2023}.

\end{abstract}

\section{Introduction}
Landslide is a serious natural disasters~\cite{lu2022regional}, and large-scale landslides threaten the local ecology and property. An important part of landslide prevention and control is predicting its displacement. Landslide surface displacement data itself is a kind of time series data, which has distinct characteristics, such as being time-sensitive, structural, and almost no update operation~\cite{Long2022}. Time series forecasting modelling is mainly aimed at its timeliness and structure, trying to give a reasonable description of its change trend, period and other characteristics, to achieve effective forecasting~\cite{Holt2004}. Traditional landslide early warning model mostly relies only on the knowledge background of land disasters, due to the scarcity and variability of landslide surface displacement evolution parameters and the complexity of the external environment, the physical model~\cite{Zou2020} can reveal its evolution mechanism, but it is difficult to obtain accurate prediction effect.

In recent years, variational modal decomposition (VMD)~\cite{Dragomiretskiy2013Variational} has been widely used in time series data decomposition and prediction because of its ability to divide series data into different sub-series with a clear physical meaning based on frequency. Zhang et al.~\cite{zhu2019Carbon} combined VMD with the optimal combination model for the carbon price prediction tasks and used VMD to complete the decomposition of surface displacement data. Due to its data-driven scalability and powerful ability to fit independent and identically distributed data, the use of machine learning(ML)-based and deep learning(DL)-based methods to predict time series data has become a relatively mature idea, and the methods that have performed better in previous tasks include XGBoost~\cite{Chen2016Xgboost}, LSTM~\cite{Hochreiter1997Long}, Prophet~\cite{Taylor2018Forecasting}, support vector regression (SVR)~\cite{Sanchez2003Advanced}, etc. Due to the particularity of geological geomorphology, it is often not feasible to use the same mathematical modelling methods in different regions. In addition, the data used to train the model is likely to not satisfy the hidden assumptions in the model resulting in inconsistency and ultimately unsatisfactory prediction performance.

The key direction of landslide surface displacement prediction is to improve the existing models according to the unique characteristics of each region, and then form a new model that adapts to the characteristics of the data. In this paper, we propose a new prediction framework combining traditional statistical ideas with ML/DL-based time series prediction models, named VMD-SegSigmoid-XGBoost-ClusterLSTM (VSXC-LSTM) for landslide surface displacement time series prediction. The main contributions of this paper are as follows:
\begin{enumerate}[$\bullet$]
\item In this paper, we propose a SegSigmoid-XGBoost-ClusterLSTM (VSXC-LSTM) landslide surface displacement time series data prediction framework based on variational modal decomposition(VMD), which is suitable for the characteristics of obvious change trend and abnormal fluctuation of landslide body surface displacement data.
\item Different from existing methods, we perform nonparametric tests on each decomposed subsequence during model training to verify its property, which ensures the consistency of the training data and model assumptions.
\item We propose a sub-model for modelling irregular subsequences obtained after the modal decomposition of time series data which is named ClusterLSTM.
\item Extensive experiments on real-world datasets show our method has good precision and generalization ability. Our VSXC-LSTM framework can provide reliable research materials for experts in the field to study the changes and development of landslides, to achieve effective warning of landslide disasters.
\end{enumerate}

\section{Related work}
Due to the variability of landslide surface displacement evolution parameters and the complexity of the external environment, the physical model~\cite{Zou2020} can reveal its evolution mechanism, but it is difficult to obtain an accurate prediction effect.  Holt~\cite{Holt2004} applied the double exponential smoothing (DES) method to trend displacement prediction. This method is simple in operation, and has good prediction performance for some specific trend sequences. Cao et al.~\cite{Cao2016} proposed an extreme learning machine (ELM) method, which takes control factors into account in the landslide surface displacement prediction model. Based on the dynamic characteristics of landslides, Xu et al.~\cite{shiluo2018} used the dynamic model long and short-term memory neural network (LSTM) to predict the cumulative displacement, and the prediction result was more accurate than that of the static model  support vector regression (SVR). Huang et al.~\cite{huang2021} used a novel recursive neural network to predict the dynamic response of slope, this model is suitable for the case of a large amount of data, and the prediction error is small. In addition, Li et al.~\cite{Li2018Robust} proposed a LSTM algorithm based on clustering ideas applied to temporal series anomaly detection, which also has application value in the prediction direction of introducing clustering ideas into LSTM modelling. Krishnan et al.~\cite{Krishnan2015Deep} uses deep Kalman filtering for counterfactual reasoning, showing the excellent filtering ability of the Kalman filter algorithm. Cong et al.~\cite{Cong2020Anomaly} applies Kalman filtering to monitoring data anomaly detection tasks. Its excellent experimental results prove the feasibility of applying the Kalman filter to anomaly detection in landslide monitoring data. In terms of parameter optimization, Cui et al.~\cite{Cui2022Optimization} proposed that based on spline curve and nested least square support vector regression (LS-SVR) functional parameter optimization, and the stable control of parameter optimization value can be achieved under actual conditions. Traditional ML and DL algorithms need to meet certain conditions to get good prediction results, but in reality, there are often cases where the data do not satisfy the implicit model assumptions, resulting in poor robustness of the model prediction results. However,compared with the existing single-model ML/DL methods, our framework has good robustness while achieving higher precision.

\section{Framework}
\begin{figure}[ht]
\centering
{\includegraphics[scale=0.30]{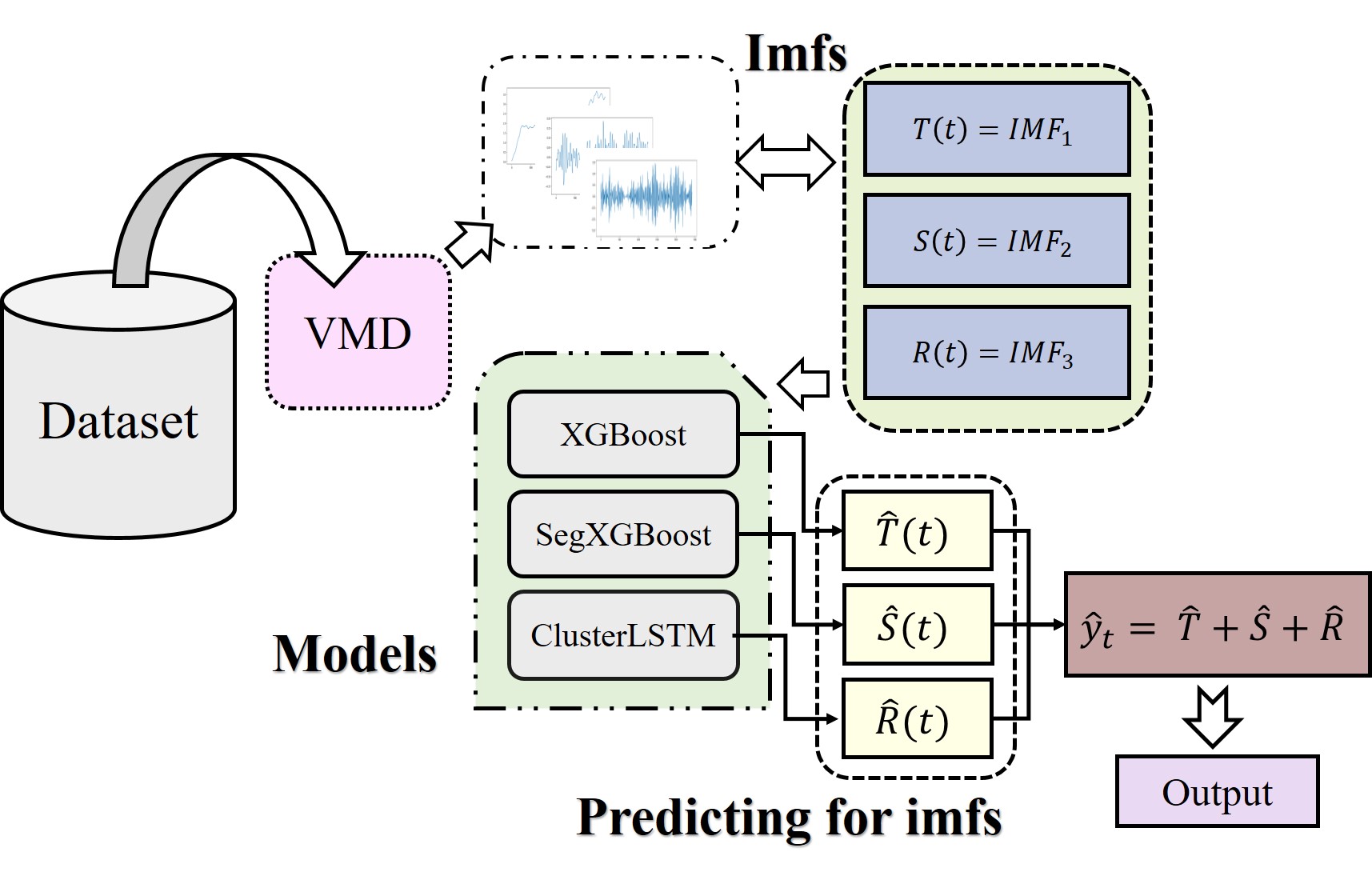}}\quad
\caption{SegSigmoid-XGBoost-ClusterLSTM model based on VMD}
\label{fig1}
\end{figure}

Our goal is to achieve better predictions by integrating models with excellent performance. This paper we propose a SegSigmoid-XGBoost-ClusterLSTM framework, as shown in Fig.~\ref{fig1}, which mainly applies modal decomposition and ensemble learning ideas. Firstly, VMD~\cite{Dragomiretskiy2013Variational} is used to decompose the time series data into three subseries, namely T (trend term), S (period term), R (residual term). Secondly, three subsequence curves are fitted using suitable models (SegSigmoid, XGBoost, ClusterLSTM), respectively. Finally, the three subseries are combined into a prediction sequence and the final prediction result is output.

\subsection{VMD decomposition based on genetic algorithm parameter}
The decomposition of the processed displacement data using the VMD algorithm can obtain more physically meaningful time series components. However, the hyperparameter penalty factor $\alpha$ and ascending step $\tau$ in the VMD algorithm will have a greater impact on the decomposition effect, to avoid the modelling bias introduced by artificially determining the hyperparameters, we use the genetic algorithm (GA) to determine a set of hyperparameter $\alpha$ that make the decomposition effect optimal by iteratively optimizing the fitness function $\tau$. It provides a general framework for solving complex system problems, which does not depend on the specific domain of the problem and has strong robustness.

Completely non-recursive VMD was proposed by Konstantin et al.~\cite{Dragomiretskiy2013Variational}. In this paper, this method is mainly used to perform modal decomposition of the original surface displacement sequence. Considering interpretability and enforceability, we prescribe the number of decompositions $K=3$ without losing a large amount of information, which are trend term, period term and residual term, respectively. The detailed steps of VMD are shown in Algorithm 1, where $u_k$ is the centre frequency of the single-component amplitude modulation signal; $\omega_k$ is the centre frequency of the single-component FM signal, $\lambda_k$ is the Lagrangian multiplier, $n$ is the number of iterations, $k$ is the number of subsequences, and $\epsilon$ is the tolerance of the convergence criterion.
\begin{algorithm}
\caption{Variational Modal Decomposition}
\label{alg:vmd}
\begin{algorithmic}[1]
\Require Raw surface displacement data y
\Ensure Trend term subseries T, S, R
\State \textbf{Initialize} $\{ u_k^1  \}$,$ \{ \omega^1_k \}$,$\lambda_k^1$,n=0,k=1,$\epsilon$=$10^{-7}$
		\While {$k \textless 3 $}
			\State $n=1$\\
			\textbf{update} $u_k^{n+1},\omega^{n+1}_k$
			\State \textbf{update} $\lambda_k $
			\If{$\sum_k \frac{||u_k^{n+1}-u_k^n||^2}{||u_k^n||_2^2}<\epsilon$}
			\State $k=k+1$
			\Else
			\State break
			\EndIf
		\EndWhile
\end{algorithmic}
\end{algorithm}

\subsection{SegSigmoid}
The SegSigmoid model is an extension of the traditional logistic regression model, which will be applied to the curve fitting and forecasting tasks of the logistic regression model used for classification tasks, and optimize the prediction effect with appropriate parameters according to different characteristics of time periods of the time series. The essence of logistic regression lies in the Sigmoid function, which is defined as follows:

\begin{align}
g(z) = \frac{1}{1+e^{-z}}
\end{align}
The characteristic of this function is that the value on the real axis domain is in the interval (0,1] and is not sensitive to maximum and minimum values. The expression of the piecewise logistic regression model is such as the formula (2):
\begin{align}
g(t) = \frac{C(t)}{1+exp(-(k+{\alpha(t)}^t\delta)(t-(m+{\alpha(t)}^T\gamma))}
\end{align}
Inspired by the Prophet algorithm, we propose the SegSigmoid algorithm and introduce anomaly detection ideas of student-based residuals~\cite{Hoaglin1978The} to obtain change point location information.

The student residual can be used to detect outliers , and calculating the residuals such as Equ (3) and (4), where $y_i$, $\hat{y}_i$ is the true value of the sequence and the predicted value of the series, respectively. Then, the studentized residual elimination dimensional differences are proposed and the Hat matrix for adjusting the sum of squares of the series residuals in (5) is introduced such as Equ (6), where n is the sample size, $X$ is the sequence dataset matrix, and $h_{ii}$ is the diagonal element of the Hat matrix. The formula is defined as follows:

\begin{align}
& \hat{y}_i = \sum_{p=0}^{N}a_p x^p_i \\
& r_i = y_i - \hat{y}_i
\end{align}
\begin{align}
& t_i = (n-p-1)\frac{r_i}{SSE(1-h_{ii})-r_i^2}\\
& H= X(X^TX)^{-1}X^T
\end{align}
We define the Bonferroni (BC) critical value to establish a suitable confidence interval in Equ (7) and (8), where $\alpha$ is the significance level, often set to 0.05, the Bonferroni critical value can be scaled equimetrically using the correction factor $\beta$ ($\beta=1/6$). The formula is defined as follows:

\begin{align}
& BC=t(1-\frac{\alpha}{2n};n-p-1)\\
& -BC(\alpha =0.05) \textless x \textless BC(\alpha =0.05)
\end{align}

\subsection{ClusterLSTM}
LSTM \cite{Hochreiter1997Long} is a long-short-term memory artificial neural network, which is developed from recurrent neural networks. The proposed ClusterLSTM introduces the idea of clustering on the basis of LSTM, aiming to solve the problem that it is difficult to use a single LSTM fitting with large differences in the scale of different window series. In addition, K-means algorithm is used to realize the clustering of residual term time series data, and shallow LSTMs are established separately for each class to improve the prediction performance.

\begin{figure}[ht]
\centering
{\includegraphics[scale=0.20]{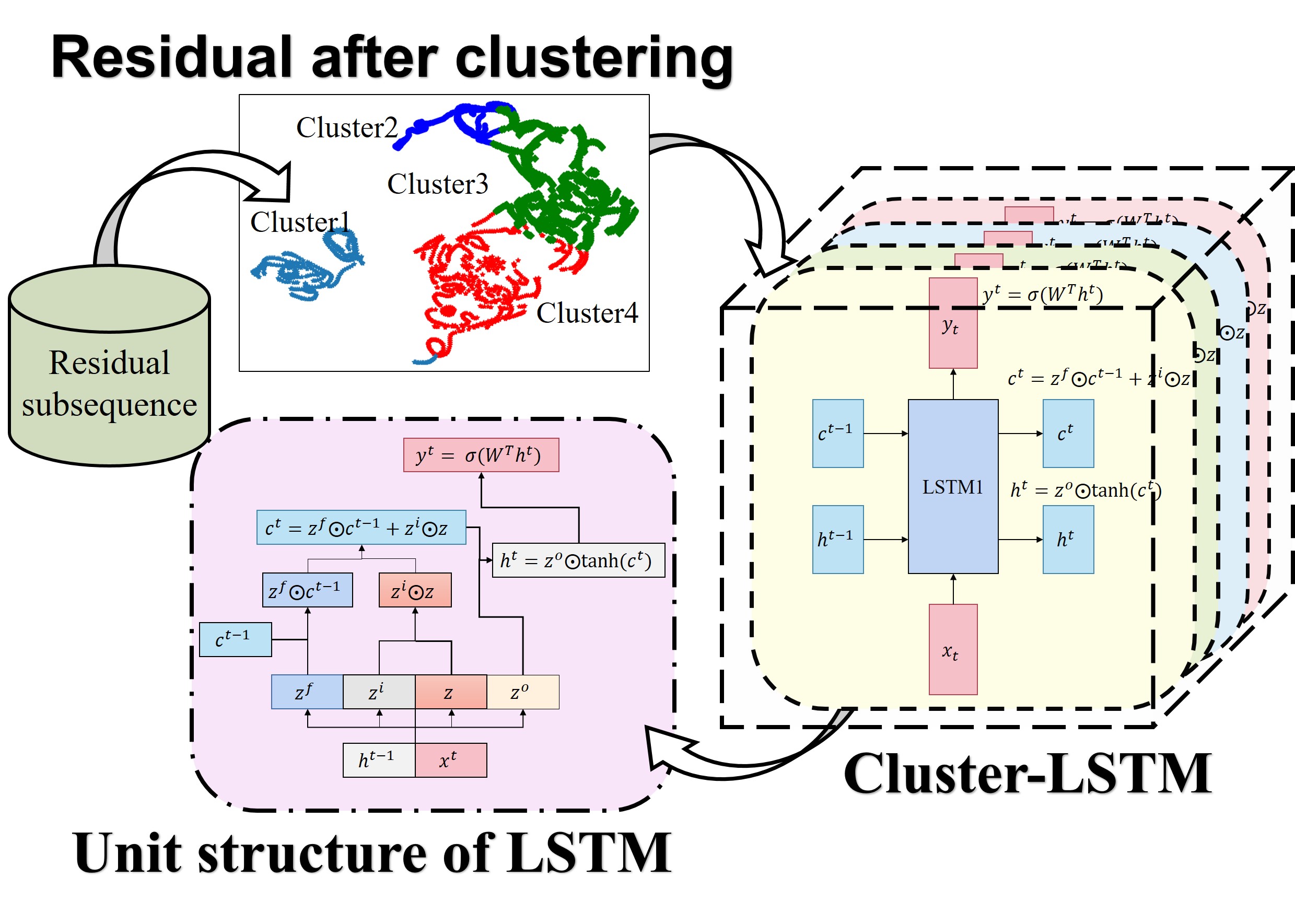}}\quad
\caption{The framework of our ClusterLSTM}
\label{fig2}
\end{figure}

Specifically, the residual term series obtained from the VMD is first dealt with a white noise test. Then, $K$-LSTM models with the same structures are initialized, and the residual items of all decomposed training samples within a batch are clustered during the training process, with the number of cluster centers being $K$, to obtain the partitioned sub-dataset $\left \{ C_{1}^{res},\cdots,C_{K}^{res} \right \} $. Then the parameters $\theta_k$ of the $k$-th LSTM model are updated as follows:

\begin{equation}
    \theta_k^{t+1} \leftarrow \theta_k^t-
\alpha\left[\sum_{i=1}^{\left | C_k^{res} \right | }
\nabla_{\theta_k} Loss\left(R_{i,k},\hat{R}_{i,k}\right)\right]
\end{equation}
where $\theta_{k}$ is the parameters of the $k$-th LSTM model, $\left | C_k^{res} \right |$ is the size of the $k$-th sub-dataset, $\alpha$ is learning rate, $R_{i,k}\in  C_k^{res} $ is the groundtruth residual term in the $k$-th sub-dataset, $\hat{R}_{i,k}$ is the output of the $k$-th LSTM model. The principle of ClusterLSTM technology is shown in Fig.~\ref{fig2}.

\section{Modeling and Results}
\subsection{Dataset introduction and data preprocessing}
The dataset used in this paper is from the relative displacement data recorded by sensors from December 12, 2020, to March 21, 2021, from the N $27\degree13'57''$ and E $109\degree42'05''$ monitoring points in the landslide area of the Niutang Pass Formation in Dashajie Village, Heliao Township, Zhijiang County. Specifically, a total of 2426 pieces of time series data recorded in hours were divided by a threshold of 0.9, the first 2184 pieces of data were used for the training of each machine learning model, and the last 242 pieces of data were used as a test set to verify the accuracy of the model and performed 10-fold cross-validation.

\begin{figure}[h!t]
\centering
\subfigure[Raw data vs. filtered data]{\includegraphics[scale=0.135]{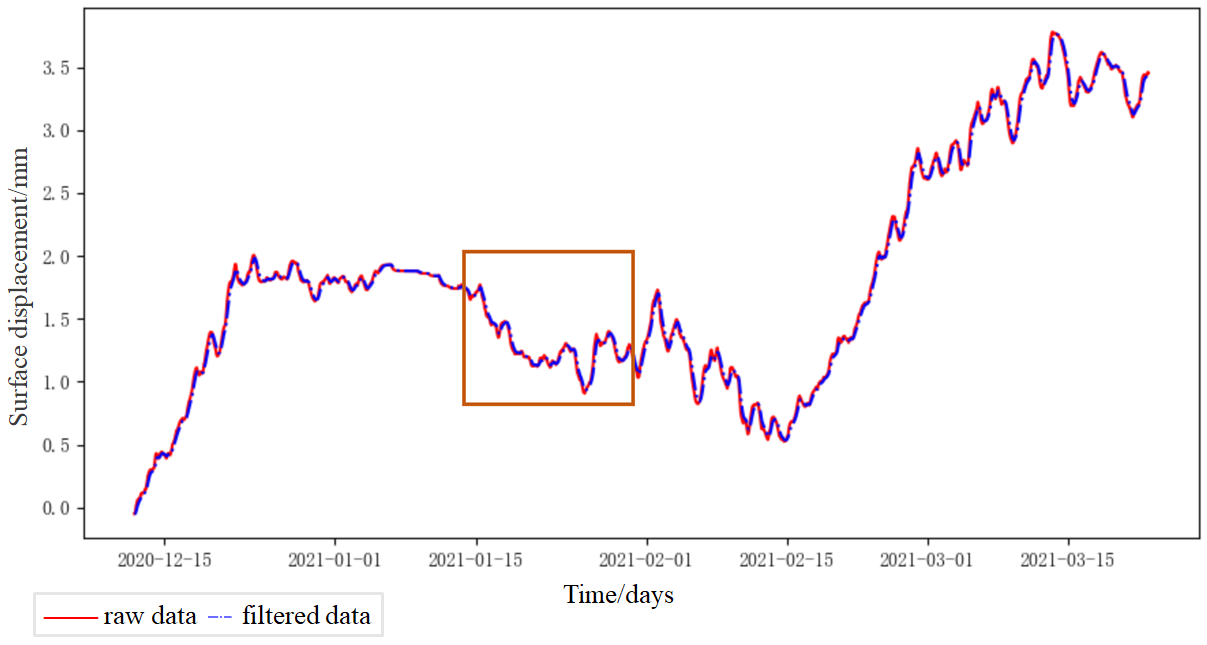}}\quad
\subfigure[Filter effect local thumbnails]{\includegraphics[scale=0.135]{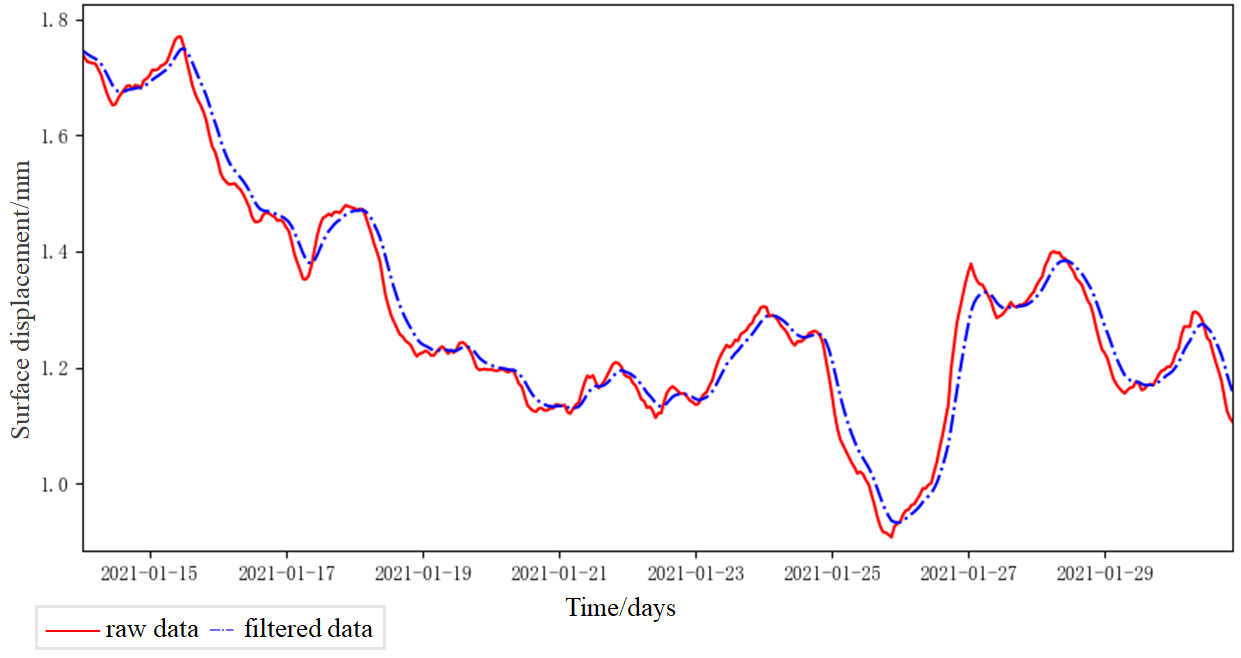}}
\caption{Sensor raw data and Kalman filtered processed data}
\label{fig3}
\end{figure}

The sensor has measurement errors and process noise when recording landslide motion, so the original displacement data contains many invalid information and interference factors. Before building the model, we first used Kalman-Filter~\cite{Welch1995An} to smooth the noise filtering of the raw data. So we set the variance of the process noise $w$ $\sim$ $N(0, Q)$ to 1, due to the degradation of the measurement accuracy due to sensor ageing, so the variance of the measurement error $v$ $\sim$ $N(0, R)$ is set to 16, and the original data and the filtered data curve are shown in Fig.~\ref{fig3}. From Fig.~\ref{fig3}-a, we find the Kalman filter retains the characteristics of the original curve to the greatest extent, and does not change the period and trend of the original curve. Observing Fig.~\ref{fig3}-b, it can be concluded that the displacement data processed by the Kalman filter has a smoother change trend in the data and shows a more obvious change regularity than the original data.

\subsection{Evaluation indicators}
In order to accurately evaluate the predictive performance of our designed model, the evaluation index is Root Mean Square Error (RMSE) and Mean Absolute Percentage Error (MAPE) as shown in Equation (10) and (11). Where $N$ is the total sample size of the prediction, and $d_i$ and $\hat{d}_i$ are the true and predicted values of sample$_i$, respectively.	The formula is defined as follows:

\begin{align}
& RMSE=\sqrt{\frac{1}{N}\sum_{i=1}^{N}(\hat{d}_i-d_i)^2  }; \\
& MAPE= \frac{1}{N}\sum_{i=1}^{N}\left |\frac{\hat{d}_i-d_i}{d_i}  \right |;
\end{align}

\subsection{VMD decomposition}
Before we use genetic algorithm (GA), we first use VMD to obtain the hyperparameter $\alpha$. Specifically, we specify the number of components after decomposition $K=3$, with the reconfiguration mean squared error $(\tilde{y}-y)^2/n$ of the decomposition sequence $\tilde{y}=imf_1+imf_2+imf_3$ as the fitness function of the GA: the number of individuals in the population $m = 50$, the number of iterations $niter = 100$, the cross probability is 0.7, and the variation probability is 0.1. Finally, we get the optimal hyperparameter $\alpha=13.625,\tau=0.99877$.
	
\begin{figure}[h!t]
\centering
\subfigure[Trend term subsequence T]{\includegraphics[scale=0.09]{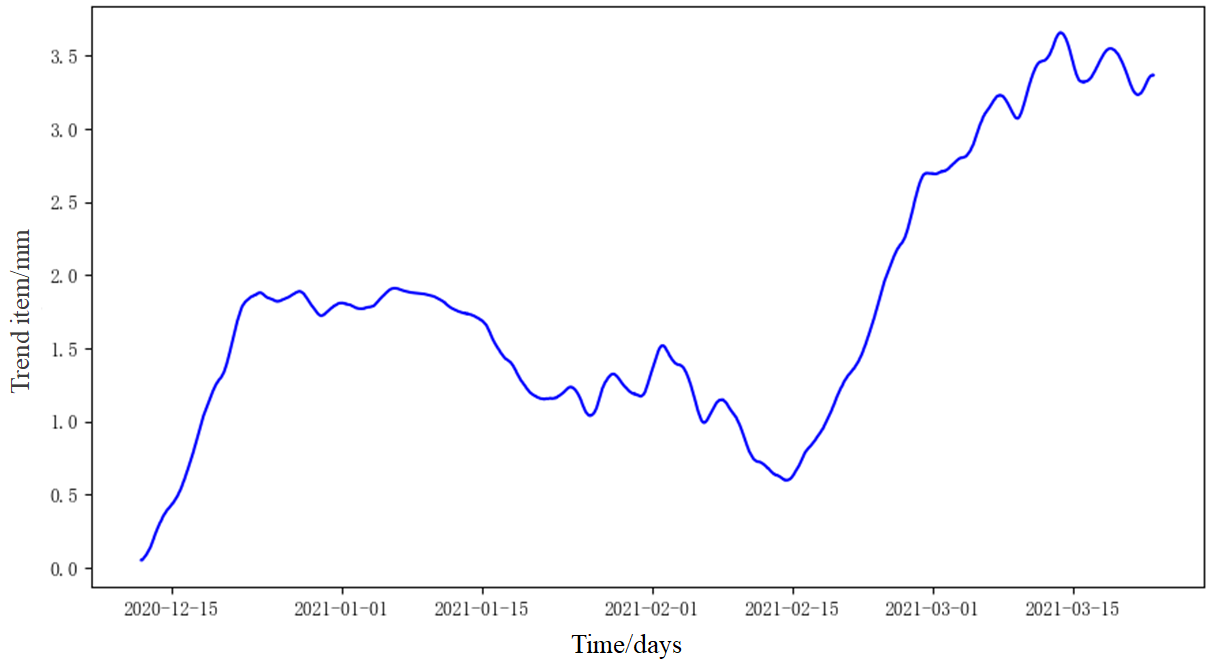}}\quad
\subfigure[Periodic term subsequence S]{\includegraphics[scale=0.09]{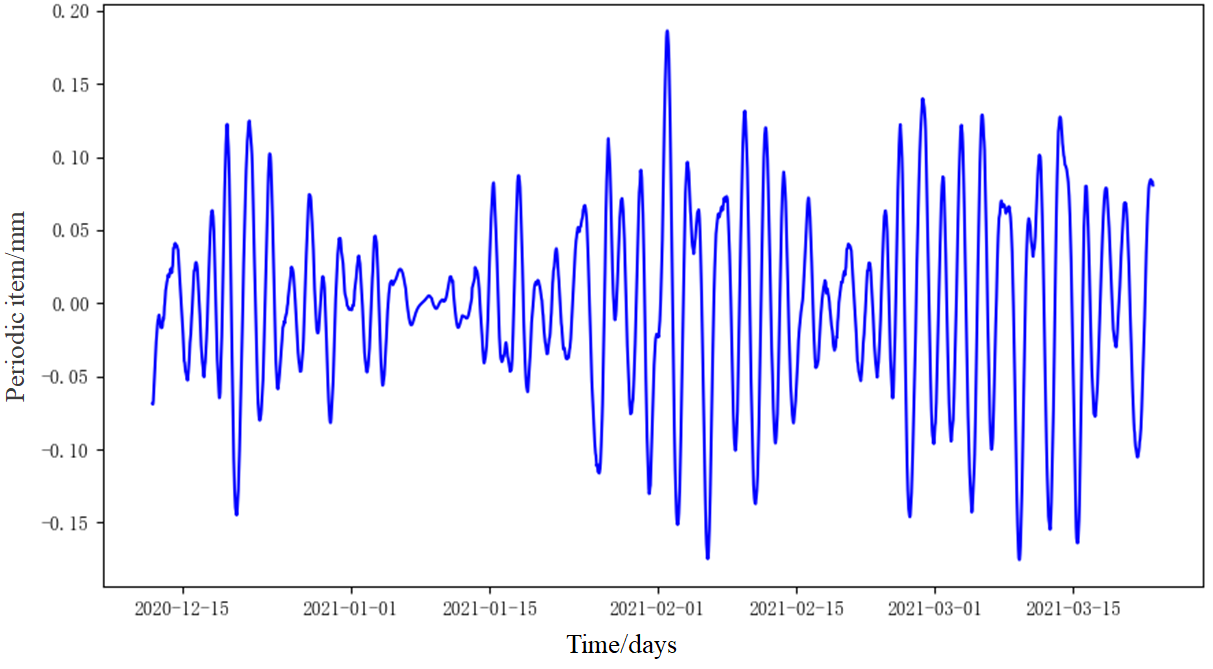}}\quad
\subfigure[Residuals term subsequence R]{\includegraphics[scale=0.09]{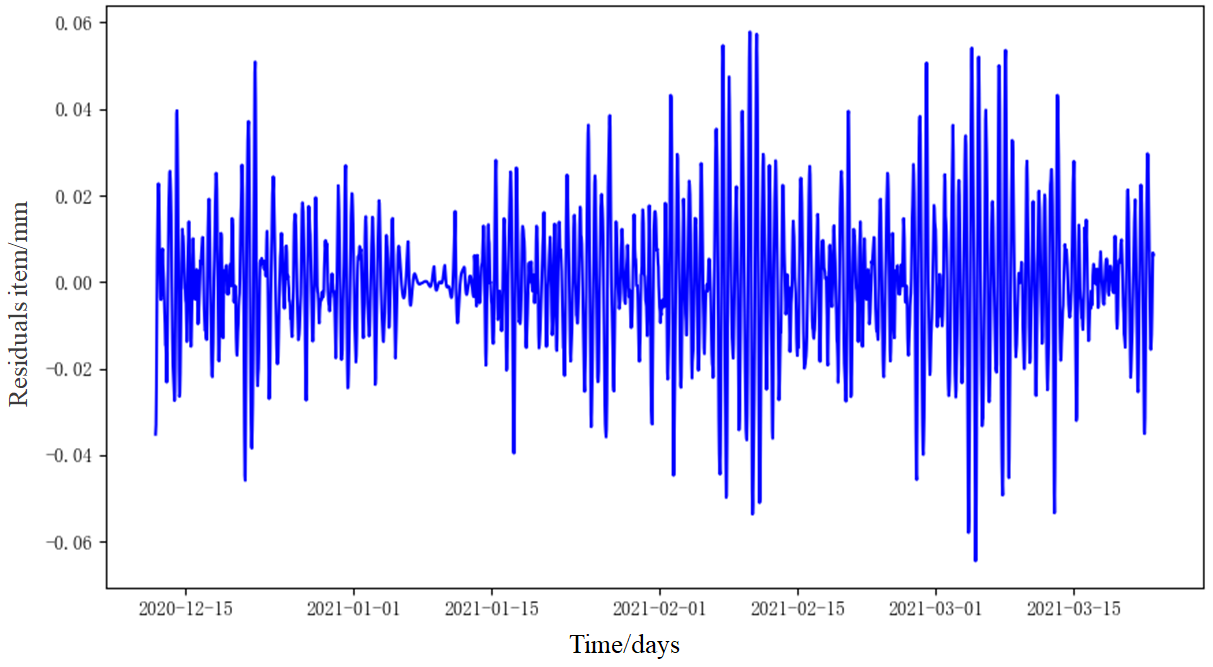}}
\caption{Three subsequences after VMD decomposition}
\label{fig4}
\end{figure}

The components decomposed by VMD are shown in Fig.~\ref{fig4}. The first eigenmode function $imf_1$ after decomposition is shown in Fig.~\ref{fig4}-a, which is smoother than the original sequence y that named trend term T; the second subsequence is shown in Fig.~\ref{fig4}-b, exhibits a strong periodicity that named periodic term S; the last subsequence is shown in Fig.~\ref{fig4}-c, and it is difficult to observe a more obvious law from the figure, and it is named residual term R.

We use the VMD algorithm to decompose the original sequence into three subsequences, namely: $y_t=T+S+R+e$, where $e$ is the error caused by VMD, this error has been minimized by optimizing the VMD algorithm parameters, and is ignored here. By establishing a time series forecasting model for three subseries, $\hat{T}$,$\hat{S}$,$\hat{R}$ are obtained, and then the final predicted value $\hat{y}$ is obtained.

\subsection{Trend term, periodic term, residual term test and model establishment}
\subsubsection*{(a)Trend term testing and model building.}
\begin{figure}[h!t]
\centering
{\includegraphics[scale=0.7]{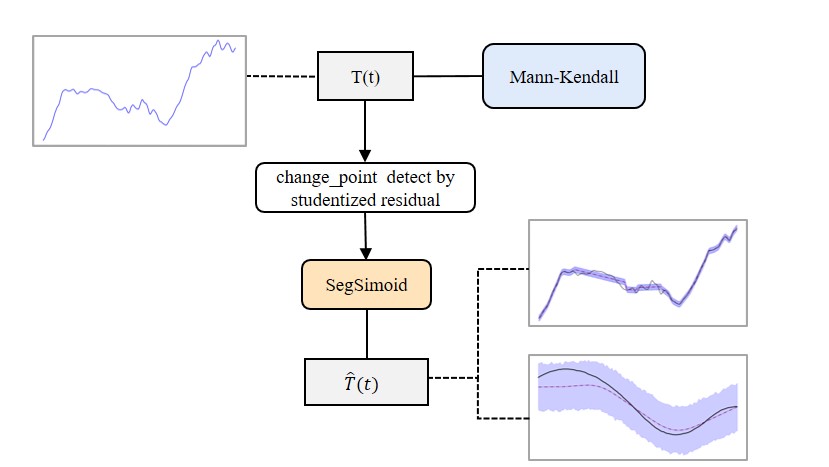}}\quad
\caption{Trend term testing and predictive modelling process}
\label{fig5}
\end{figure}

For the trend term subseries, we first perform the Mann-Kendall test on the sequence. Based on the actual prediction effect, we decided to use SegSigmoid as the trend term prediction model. The trend item modelling process is shown in Fig.~\ref{fig5}.

\begin{figure}[ht]
\centering
{\includegraphics[scale=0.20]{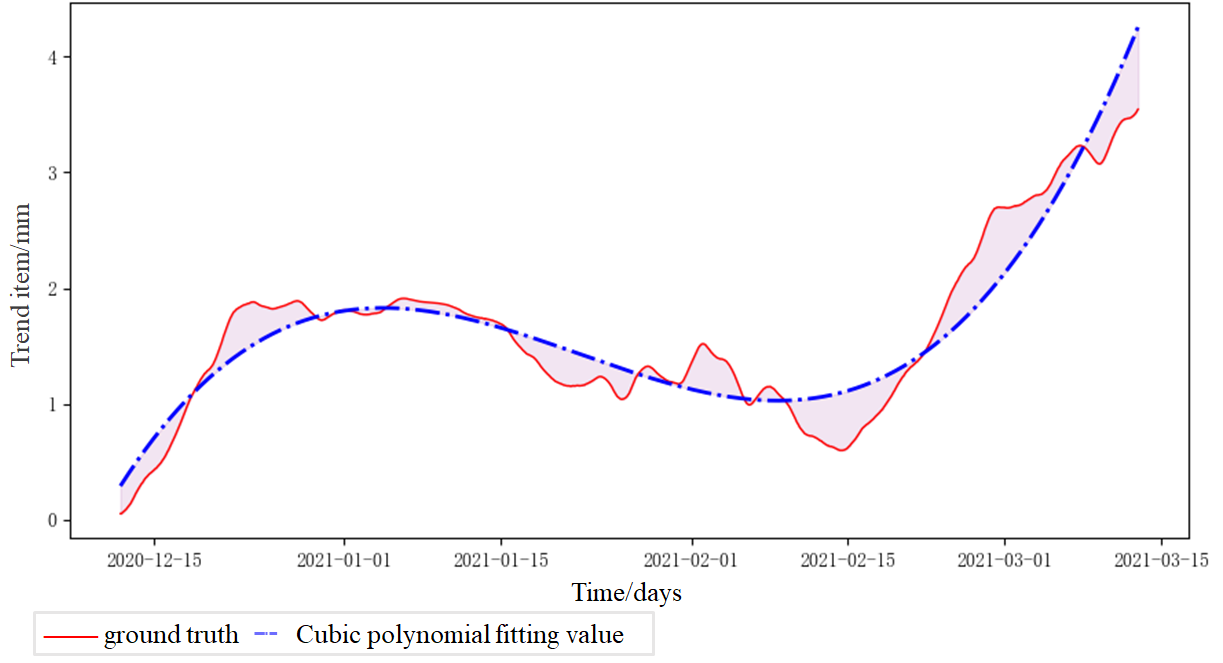}}\quad
\caption{Cubic polynomial fits the basic shape of a trend term}
\label{fig6}
\end{figure}

First, the Mann-Kendall tendency test in the nonparametric test is used to determine the trendiness of the target series, and the specific results are shown in Table~\ref{tab1}. So we have sufficient evidence to reject the null hypothesis, that the series has a trend, and can be modelled and predicted on this basis. It is observed that the change of the trend term T approximately follows the change law of the cubic polynomial curve $y_t=at^3+bt^2+ct+d$, as shown in Fig.~\ref{fig6}.

\begin{table}[ht]
\centering
\caption{Results of the Mann-Kendall test}
\label{tab1}
\tabcolsep 7pt
\begin{tabular}{lcccc}
\toprule
\multirow{1}{*}{$\text{Mann-Kendall}$} & original hypothesis & Z statistic & p-value & conclusion \\
\midrule
\multirow{1}{*}{$\text{Results}$} & no trend of T & 22.059 & 0.000 & Rejection hypothesis \\
\bottomrule
\end{tabular}
\end{table}

We fit the SegSigmoid function on the training set and calculate the studentized residuals at each moment after cubic polynomial fitting to mark outlier points and set the set of outliers to the mutation points of the SegSigmoid function. After calculation, the number of mutation points on the training set is 1146, and the growth rate changes $\delta{j}$ at the mutation point $s_j$ obeys Laplace $Laplace(0,\tau)$, where $\tau$ is 0.5, and the distribution range of mutation points on the training set is 0.95, only change points  on the first 95\% of the training set.

\subsubsection*{(b)Periodic term inspection and model establishment.}

\begin{figure}[h!t]
\centering
\subfigure[Periodic term testing and prediction process]{\includegraphics[scale=0.40]{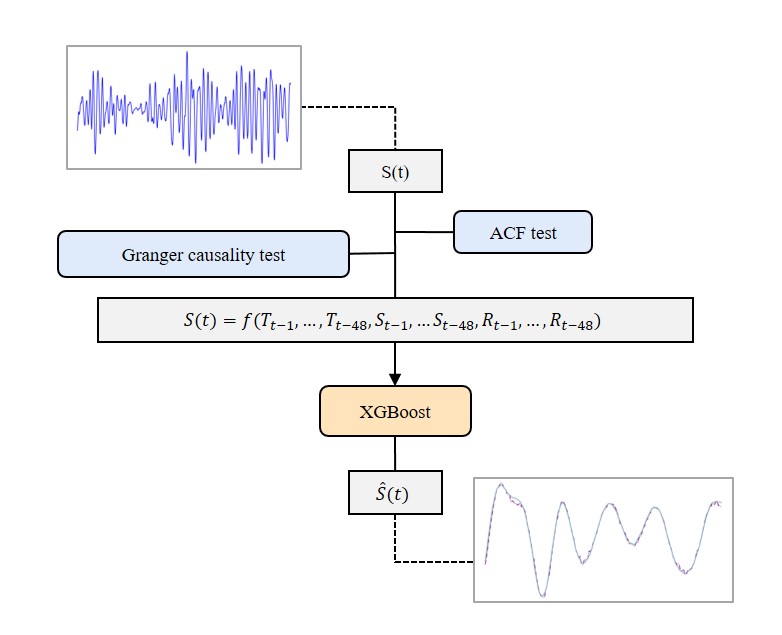}}\quad
\subfigure[Periodic term autocorrelation test]{\includegraphics[scale=0.4]{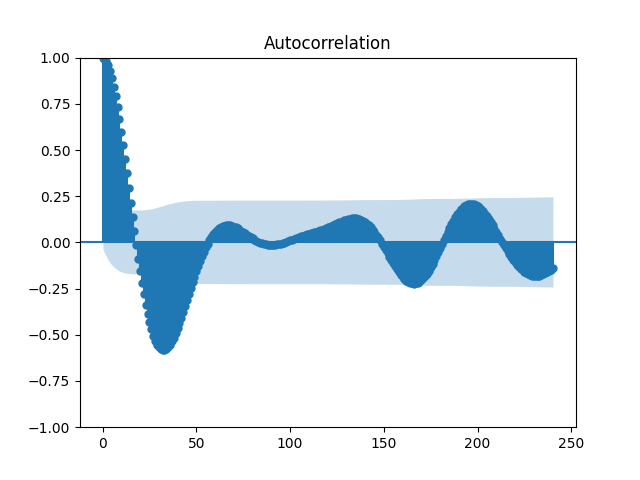}}\quad
\caption{Periodic term}
\label{fig7}
\end{figure}

Periodic term modelling is different from trend term, we first perform an autocorrelation test on periodic term, and then perform the Granger causality test for the proposed hypothesis. Finally, we determine the final prediction model of the periodic term as XGBoost according to the performance of the prediction set. The process of checking and predicting periodic terms is shown in Fig.~\ref{fig7}-a. To determine the lagged dependence of terms in the periodic term $S$, the autocorrelation function (ACF) of the periodic term $S$ is first calculated. From Fig.~\ref{fig7}-b, the ACF curve disappears at the lag 48 period, which indicates that the autocorrelation function is truncated at the lag 48 periods, which means that relative displacement at a certain moment is most affected by the displacement data 48 hours ago. We assume that the period term $S$ is simultaneously affected by the trend term $T$, the period term $S$, the residual term $R$, and the relative displacement $y$ with lags from periods 1 to 48, so there is a prediction function $f$ that maps the time series component of the lag to the period term $S_t$ of the current period:
\begin{align}
S_t = f(T_{t-1},\cdots T_{t-48},S_{t-1},\cdots S_{t-48},R_{t-1},\cdots R_{t-48},y_{t-1},\cdots y_{t-48})
\end{align}
To verify the correctness of this hypothesis, we use the Granger causality test to determine whether the causal relationship between the period term $S$, $T$, $R$; and the results of the Granger causal test with a lag of 48 periods are shown in Table~\ref{tab2}. We can see the period term prediction model is statistically significant.

\begin{table}[h!t]
\centering
\caption{Results of Granger causal test under 48-order hysteresis}
\label{tab2}
\tabcolsep 3pt
\begin{tabular}{lcccc}
\toprule
\multirow{1}{*}{$\text{Test object}$} & Original hypothesis & F statistic & P-value & Conclusion \\
\midrule
\multirow{1}{*}{$\text{Relative displacement$y$}$} & Y is not S's Granger reason & 180825.2560 & p=0.0000 & Reject  \\
\multirow{1}{*}{$\text{Trend item $T$}$} & T is not S's Granger reason & 26297.4066 & p=0.0000 & Reject  \\
\multirow{1}{*}{$\text{Residuals item $R$}$} & S is not S's Granger reason & 548946.1211 & p=0.0000 & Reject  \\
\bottomrule
\end{tabular}
\end{table}

After completing the Granger causal test, we use the XGBoost model with excellent performance in a number of similar prediction tasks as the prediction model of the periodic term S and select the same common support vector regression (SVR) model as a comparison. 

\subsubsection*{(c)Residual term test and model establishment.}
\begin{figure}[h!t]
\centering
{\includegraphics[scale=0.7]{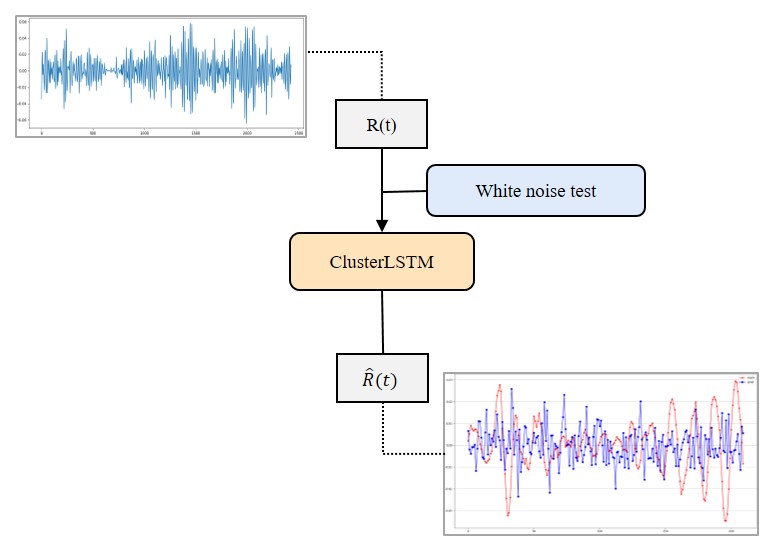}}\quad
\caption{Residual term inspection and prediction process}
\label{fig8}
\end{figure}

In view of the problems such as the difficulty of capturing the residual term law, and the suspected white noise, the white noise test is carried out on the residual term subsequence, and the prediction analysis is carried out after confirming that it is a non-white noise sequence. Experiments show that the prediction performance of LSTM directly is poor, so the ClusterLSTM model is introduced and then established separately. The process of testing and predicting residual terms is shown in Fig.~\ref{fig8} and the Ljung-Box test results are shown in Table~\ref{tab3}. After testing, it is concluded that there is still a sequence dependence in the residual term R that has not been completely extracted, so the residual term R is a non-white noise sequence.

\begin{table}[h!t]
\centering
\caption{First-order hysteresis white noise test results}
\label{tab3}
\tabcolsep 7pt
\begin{tabular}{lcccc}
\toprule
\multirow{1}{*}{$\text{original hypothesis}$} & Ljung-Box statistic & p-value & conclusion \\
\midrule
\multirow{1}{*}{$\text{R is the white noise sequence}$} & 1817.319 & 0.00 & Rejection hypothesis \\
\bottomrule
\end{tabular}
\end{table}

Since the data show the characteristics of large-scale differences in different window sequences, we consider ClusterLSTM which first use K-means clustering~\cite{Hartigan1979Algorithm} to aggregate subsequences of similar scales into K different clusters, and then train K-LSTM models with simpler structures on the data in each cluster to enhance the model's perception of data scale changes.

\begin{figure}[h!t]
\centering
{\includegraphics[scale=0.45]{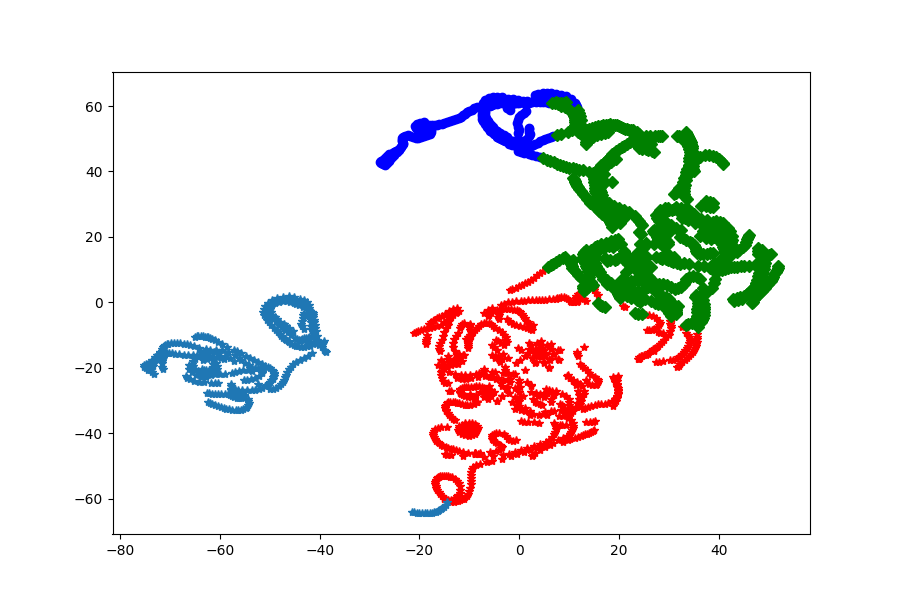}}\quad
\caption{Distribution of residual term time series samples in two-dimensional feature space}
\label{fig9}
\end{figure}

In the experiment, we use 24 as a window length to divide the entire time series into subsequences, and each subsequence as a sample. Then, the t-SNE~\cite{Van2014Accelerating} algorithm is used to convert the 24-dimensional time series samples into the 2-dimensional feature space for visual display, and the visualization effect is shown in Fig.~\ref{fig9}. From Fig.~\ref{fig9}, time series samples can be roughly divided into four categories, so we set the cluster $K=4$ in the K-means clustering algorithm. A shallow LSTM with a hidden layer number of 2 and a hidden layer neuron number of 6 is trained for the samples in each cluster.

\subsection{Series summary prediction and model validation}
\begin{figure}[h!t]
\centering
{\includegraphics[scale=0.20]{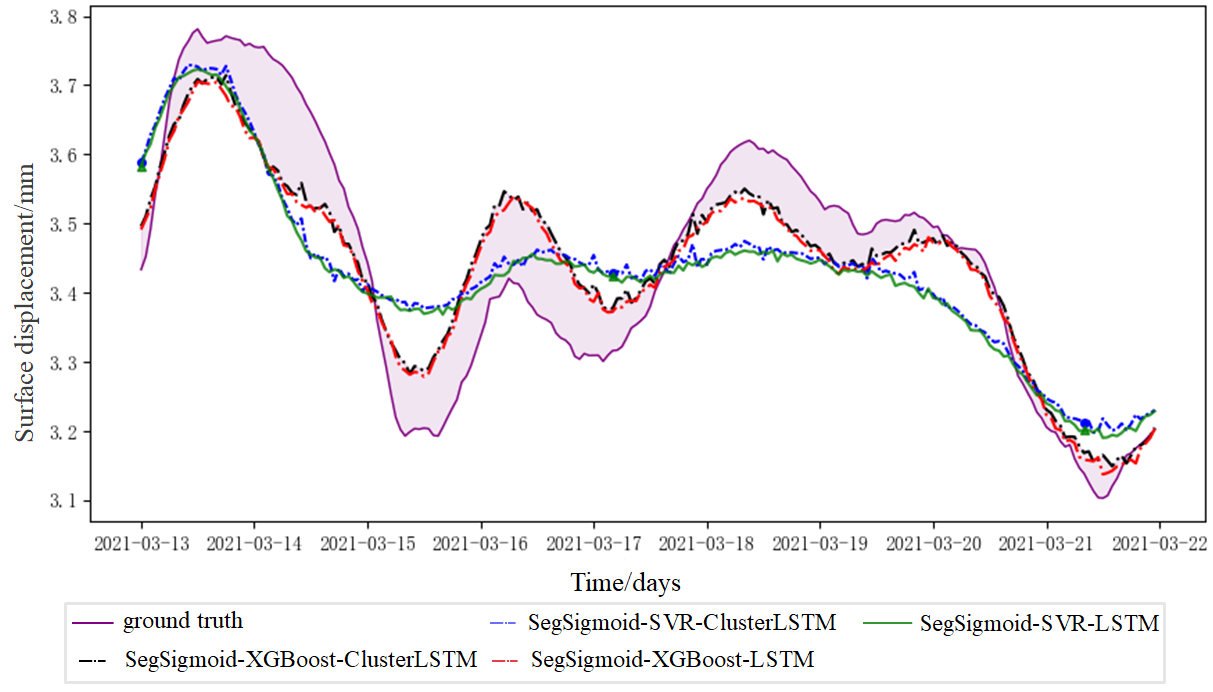}}\quad
\caption{The performance of different total displacement prediction models on the test set}
\label{fig10}
\end{figure}

According to the principle of time series, the displacements of the trend, period, and residual terms are added together to obtain the displacement prediction value $\hat{y}$, and prediction results are shown in Fig.~\ref{fig10}. It can be seen that the predicted values obtained by different forecasting methods have high similarity with the actual values, and the predictions of the overall trend are basically consistent with the actual values, but the predictions of local fluctuations by different methods have great differences, among which the prediction curve obtained by the VSXC-LSTM framework is the closest to the actual value. As shown in Table~\ref{tab4}, it show that we propose has achieved the highest prediction accuracy on the two evaluation indicators, which verifies the effectiveness of the method.

\begin{figure}[h!t]
\centering
{\includegraphics[scale=0.20]{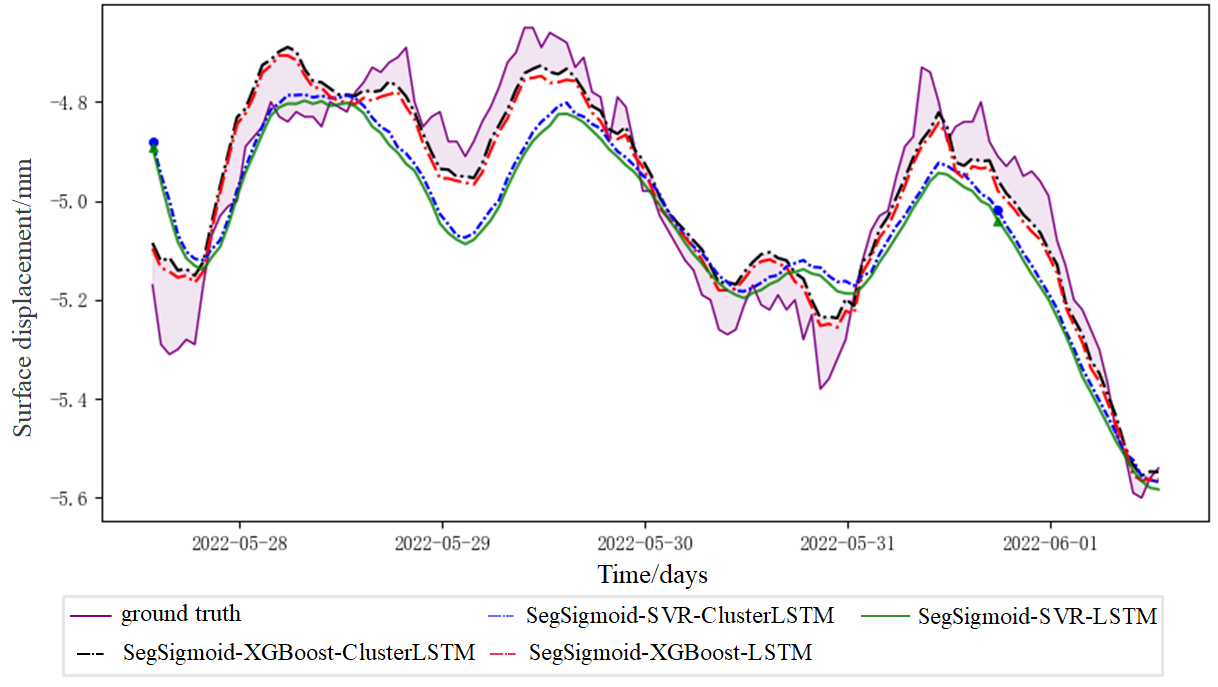}}\quad
\caption{The performance of different total displacement prediction models on validation sets}
\label{fig11}
\end{figure}

\begin{table}[h!t]
\centering
\caption{The performance of different total displacement prediction models on the test set}
\label{tab4}
\tabcolsep 9pt
\begin{tabular}{lcccc}
\toprule
\multirow{1}{*}{$\text{Predictive models}$} & RMSE & MAPE \\
\midrule
\multirow{1}{*}{$\textbf{SegSigmoid-XGBoost-ClusterLSTM}$} & $\mathbf{0.0744}$ & $\mathbf{0.0177}$ \\
\multirow{1}{*}{$\text{SegSigmoid-SVR-ClusterLSTM}$} & 0.0924 & 0.0242\\
\multirow{1}{*}{$\text{SegSigmoid-XGBoost-LSTM}$} & 0.0787 & 0.0183\\
\multirow{1}{*}{$\text{SegSigmoid-SVR-LSTM}$} & 0.0958 & 0.0242\\
\bottomrule
\end{tabular}
\end{table}
In order to further verify the effectiveness and generalization performance of our method, we use the recent relative displacements recorded by sensors from March 20, 2022, to May 28, 2022, at different monitoring points in the same region as the validation dataset, the results are shown in Fig.~\ref{fig11}, similarly, we give a pair of different method combinations on the validation set as shown in Table~\ref{tab5}. It can be seen that the prediction framework we proposed still performs well on the new dataset, especially the VSXC-LSTM framework achieves the best on both evaluation indicators. This shows that our method has high prediction accuracy and has robustness for different data sets.

\begin{table}[h!t]
\centering
\caption{The performance of different total displacement prediction models on validation sets}
\label{tab5}
\tabcolsep 9pt
\begin{tabular}{lcccc}
\toprule
\multirow{1}{*}{$\text{Predictive models}$} & RMSE & MAPE \\
\midrule
\multirow{1}{*}{$\textbf{SegSigmoid-XGBoost-ClusterLSTM}$} & $\mathbf{0.0701}$ & $\mathbf{0.0121}$ \\
\multirow{1}{*}{$\text{SegSigmoid-SVR-ClusterLSTM}$} & 0.1081 & 0.0189\\
\multirow{1}{*}{$\text{SegSigmoid-XGBoost-LSTM}$} & 0.0709 & 0.0122\\
\multirow{1}{*}{$\text{SegSigmoid-SVR-LSTM}$} & 0.1091 & 0.0189\\
\bottomrule
\end{tabular}
\end{table}

\section{Conclusion and Outlook}
We present a framework for predicting the surface displacement of landslides. Our approach uses the Kalman filter to smooth the original sequence, VMD to divide the time series into three subseries, and GA to optimize hyperparameters. We propose the SegSigmoid model for high fitting of the trend term and XGBoost to predict the periodic term. To address the insensitivity of LSTM to the data scale, we introduce ClusterLSTM. Our proposed framework achieves high prediction accuracy and outperforms other models. However, there is still room for improvement, such as reducing the number of independent prediction models and developing an end-to-end training-prediction framework based on neural networks. Additionally, expanding the proposed univariate model to a multivariate prediction model is an essential future task, as soil characteristics, structure, rainfall, and other factors significantly impact landslide displacement.

\textbf{Acknowledgements}  This study was supported by Natural Science Foundation of Hunan Province (grant number 2022JJ30673) and by the Graduate Innovation Project of Central South University (2023XQLH032, 2023ZZTS0304).

\end{document}